\pdfoutput=1

\documentclass[11pt]{article}

\usepackage[final]{acl}
\usepackage{enumitem}
\usepackage{multirow}
\usepackage{times}
\usepackage{latexsym}

\usepackage[T1]{fontenc}

\usepackage[utf8]{inputenc}
\usepackage[english]{babel}

\usepackage{microtype}
\usepackage[english]{babel}
\usepackage{inconsolata}

\usepackage{graphicx}
\title{Hybrid OCR-LLM Framework for Enterprise-Scale Document Information Extraction Under Copy-heavy Task}

\author{
  Zilong Wang \\
  Ningbo Institute of Digital Twin \\
  Eastern Institute of Technology \\
  Ningbo, Zhejiang 315200, P.R. China \\
  \texttt{zlw@idt.eitech.edu.cn}
  \And
  Xiaoyu Shen\thanks{Corresponding Author} \\
  Ningbo Institute of Digital Twin \\
  Eastern Institute of Technology \\
  Ningbo, Zhejiang 315200, P.R. China \\
  \texttt{xyshen@eitech.edu.cn}
}

\begin{document}
\maketitle
\begin{abstract}
Information extraction from copy-heavy documents, characterized by massive volumes of structurally similar content, represents a critical yet understudied challenge in enterprise document processing. We present a systematic framework that strategically combines OCR engines with Large Language Models (LLMs) to optimize the accuracy-efficiency trade-off inherent in repetitive document extraction tasks. Unlike existing approaches that pursue universal solutions, our method exploits document-specific characteristics through intelligent strategy selection. We implement and evaluate 25 configurations across three extraction paradigms (direct, replacement, and table-based) on identity documents spanning four formats (PNG, DOCX, XLSX, PDF). Through table-based extraction methods, our adaptive framework delivers outstanding results: F1=1.0 accuracy with 0.97s latency for structured documents, and F1=0.997 accuracy with 0.6 s for challenging image inputs when integrated with PaddleOCR, all while maintaining sub-second processing speeds. The 54× performance improvement compared with multimodal methods over naive approaches, coupled with format-aware routing, enables processing of heterogeneous document streams at production scale. Beyond the specific application to identity extraction, this work establishes a general principle: the repetitive nature of copy-heavy tasks can be transformed from a computational burden into an optimization opportunity through structure-aware method selection.
\end{abstract}

\section{Introduction}

\begin{figure*}[htb]
  \centering
  \includegraphics[width=0.9\linewidth]{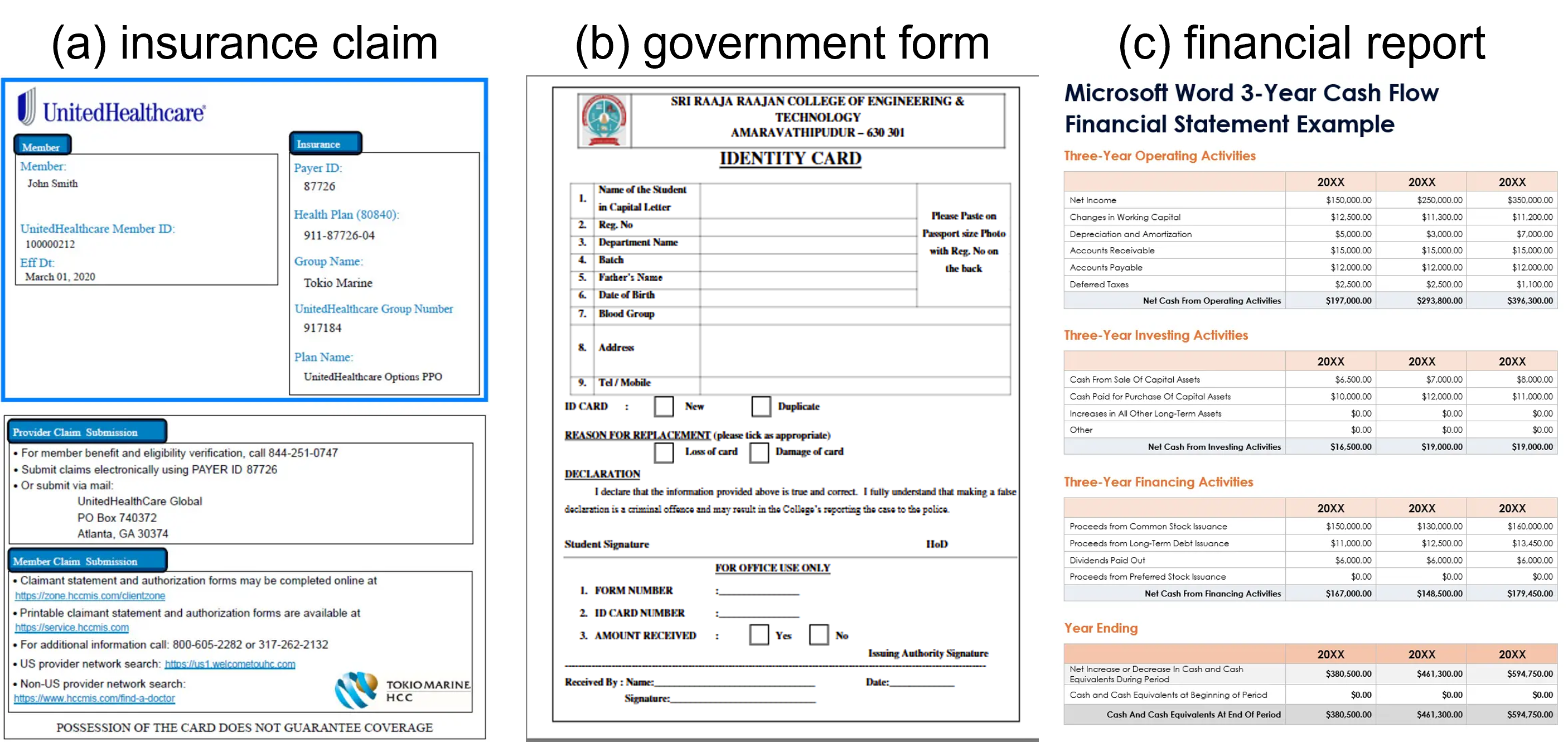}
  \caption {Document examples for copy-heavy tasks: (a) insurance claim; (b) government form; (c) financial report.}
  \label{fig:examples}
\end{figure*}

Copy-heavy tasks, characterized by large-scale processing of highly repetitive and template-based documents, pose persistent challenges in enterprise environments. From insurance claims and government forms to financial reports and identity documents, as shown in Figure~\ref{fig:examples}, enterprises routinely extract structured information from millions of similar documents daily~\cite{gagie2017document,navarro2019document,cobas2019fast}. While the structural redundancy offers potential for optimization, it also exacerbates issues of computational inefficiency, error propagation, and system brittleness.

Traditional extraction systems rely heavily on rule-based or template-specific configurations~\cite{majumder2020representation,xu2020layoutlm}. These approaches often perform well under controlled conditions but degrade rapidly with minor format changes, limiting their scalability across diverse enterprise settings~\cite{gunel2022data,zhang2024document}. Moreover, quality assurance remains difficult without labeled benchmarks~\cite{seitl2024assessing}, and integration with downstream systems is often ad hoc and fragile~\cite{tang2021making}.

Large Language Models (LLMs) offer new possibilities for zero-shot and instruction-based document understanding. However, their practical use for structured extraction remains limited by high latency, hallucination risks~\cite{li2024dawn}, and inefficiencies in handling repetitive or low-variance content. In copy-heavy tasks, latency is especially pronounced because generative models must decode output token by token; what an OCR/post-processing stack can copy in (near) constant time becomes hundreds or thousands of sequential generations. This generation-first workflow wastes time and budget on text that could be copied verbatim and increases exposure to stochastic errors. In production pipelines that demand sub-second response times and high precision—such as identity information extraction—these limitations can lead to unacceptable failure rates~\cite{cooney2023end}.

Identity document processing exemplifies the core challenges of copy-heavy extraction. Fields such as names, dates, and ID numbers appear in consistent formats across documents, yet achieving 100\% accurate, high-throughput extraction remains elusive. Minor OCR errors or model inconsistencies can introduce systematic faults across entire datasets, highlighting the need for robust, adaptive extraction frameworks.

In this paper, we propose a fast and adaptive multi-method extraction framework that integrates traditional OCR engines with LLM-based strategies to process copy-heavy documents efficiently. Our contributions are as follows.

\begin{itemize}[itemsep=0pt,parsep=0pt,topsep=0pt]
    \item \textbf{A hybrid extraction framework} combining multiple OCR engines with four LLM-based paradigms—Direct, Replace, Table, and Multimodal—tailored to different document structures and modalities.

    \item \textbf{A diverse empirical evaluation} of 25 method combinations over documents spanning multiple formats (e.g., PNG, DOCX, XLSX, PDF), uncovering strategy-specific trade-offs in speed and accuracy.
    
    \item \textbf{A document-aware method selection strategy} that achieves perfect F1 scores (1.000) with sub-second latency (0.97s average) by matching extraction methods to document characteristics.
    
    \item \textbf{Practical deployment insights} including guidance on OCR engine selection, integration challenges, and system-level design for scalable enterprise deployment.
\end{itemize}

Our work bridges the gap between LLM capabilities and production demands, providing a practical pathway for deploying high-performance extraction systems across repetitive document tasks in real-world enterprise settings.

\section{Related Work}

\textbf{Traditional Paradigms in Document Information Extraction.}
Early systems for information extraction (IE) from structured or semi-structured documents were predominantly rule-based or template-driven~\cite{appelt1999introduction,chiticariu2013rule}. These methods offered high precision and interpretability without the need for annotated training data. However, they were notoriously brittle—any minor variation in layout, field order, or formatting could break the rules, leading to maintenance-heavy pipelines that scale poorly~\cite{li2020survey}. To reduce rule authoring overhead, early work explored template induction~\cite{anick1992versioning,freitag2000machine}, and more recently, TWIX~\cite{lin2025twix} exploits redundancy across similar documents to automate extraction. Nonetheless, the lack of robustness to visual drift and document variation remains a limitation~\cite{majumder2020representation,wang2020cross}.

To address annotation costs, unsupervised techniques have been explored, such as pattern mining~\cite{etzioni2005unsupervised} and structural clustering~\cite{allahyari2017brief}, which infer field patterns across documents without labels. However, these approaches often yield noisy outputs and require significant post-processing~\cite{riedel2010modeling}. Feature-based supervised models~\cite{finkel2005incorporating,ratinov2009design} introduced stronger learning capacity, but still relied heavily on handcrafted features and domain-specific engineering, with limited generalizability to new document layouts or types~\cite{yadav2019survey}.

\textbf{Neural and Pretrained Architectures for Document Understanding.}
The advent of deep learning introduced a major shift. Sequence models like BiLSTM-CRF~\cite{lample2016neural,ma2016end,shen2017estimation} enabled end-to-end learning for token classification tasks, significantly reducing manual feature design. These were soon outpaced by pre-trained language models (PLMs)~\cite{devlin2019bert,liu2019roberta,su2022rocbert}, which provided strong contextual embeddings for text-based extraction. However, most PLMs ignore document structure, which is critical in forms, invoices, and tables.

To incorporate visual and spatial cues, layout-aware PLMs such as LayoutLM~\cite{xu2020layoutlm}, LayoutLMv2/v3~\cite{huang2022layoutlmv3}, and FormNet~\cite{lee2022formnet} combine text, layout, and visual embeddings. More sophisticated multimodal transformers like DocFormer~\cite{appalaraju2021docformer} further improve modality fusion and downstream accuracy. These models show strong performance on visually complex tasks but typically require finetuning and careful preprocessing (e.g., OCR + bounding boxes), and remain sensitive to layout noise and OCR errors.

\textbf{LLMs and Prompt-based Extraction.}
Large Language Models (LLMs) have introduced powerful new paradigms for document IE, particularly in zero-shot and few-shot settings. Prompt-based models like GPT~\cite{wang2023gpt,wei2023zero} and InstructUIE~\cite{wang2023instructuie} frame extraction tasks as natural language instruction following. Methods such as self-prompting~\cite{li2022self} and chain-of-thought reasoning~\cite{wei2022chain} improve consistency and reasoning over complex inputs. Architecturally, two key approaches emerge: (1) OCR-to-LLM pipelines like LMDX~\cite{perot2023lmdx} and DocLLM~\cite{wang2023docllm}, which linearize OCR text into prompts for token-level generation, and (2) multimodal LLMs~\cite{kim2022ocr,tang2023unifying} that directly process document images, bypassing OCR and jointly modeling layout and content.

While these models are flexible and powerful, they pose several practical challenges: inference latency, high compute cost, difficulty in debugging hallucinations~\cite{li2024dawn,zhang2023siren}, and limits in handling highly repetitive or deterministic document structures~\cite{cooney2023end}. Furthermore, most current benchmarks (e.g., FUNSD~\cite{jaume2019funsd}, DocVQA~\cite{mathew2021docvqa}) emphasize semantic diversity and visual clutter, underrepresenting production settings dominated by redundant, high-volume forms.

\textbf{Copy-heavy Extraction: A Distinct and Underexplored Setting.}
In many industrial scenarios, document extraction is not about semantic diversity but rather \textbf{structural redundancy}: processing millions of near-identical layouts (e.g., utility bills, ID cards, customs forms). We refer to this as \textbf{copy-heavy extraction}. In these settings, speed, fault tolerance, and robustness to minor OCR/model variance outweigh the need for general reasoning or layout flexibility. Even small errors can cascade into critical downstream failures in enterprise systems.

Existing LLM and multimodal models are often over-engineered for such tasks, leading to inefficiencies. Meanwhile, rule-based or template systems are fast but brittle. This tension between generality and efficiency is not well captured in most academic benchmarks, and the trade-offs in copy-heavy settings remain poorly studied~\cite{seitl2024assessing}. These tasks demand tailored strategies that optimize for high-throughput, low-latency, and graceful degradation in the presence of noise.

\textbf{Scalability Challenges and Emerging Optimizations.}
Recent work has started addressing these issues from a systems perspective. Techniques like batch prompting~\cite{cheng2023batch}, prefix sharing (BatchLLM~\cite{zheng2024batchllm}), and intermediate caching (PromptCache~\cite{gim2024prompt}, AttentionStore~\cite{zhang2023h2o}) reduce redundant computation in LLM-based pipelines, particularly when input documents are similar. Lightweight architectures such as Donut~\cite{kim2021donut} and UDOP~\cite{tang2023unifying} offer OCR-free alternatives that retain structured output formats. However, these models still suffer from trade-offs in latency, model size, and control granularity—key considerations for real-world deployments. Industrial settings also require interoperability with diverse file formats (PDF, DOCX, scanned images), graceful fallback mechanisms, and integration with legacy systems.

\textbf{Toward Modular, Hybrid, and Document-aware Pipelines.}
Copy-heavy document extraction tasks—common in industrial settings like invoices, IDs, and utility forms—prioritize speed, robustness, and consistency over general reasoning. Existing solutions either rely on brittle rules or overgeneralized LLMs, leading to inefficiencies in scalability, latency, or maintainability.

We propose a modular and hybrid pipeline that integrates OCR fusion, lightweight heuristics (e.g., field substitution, table-position extraction), and selective LLM invocation. A centralized, document-aware controller dynamically orchestrates these components based on document format, enabling fast, accurate, and scalable extraction. Unlike prior approaches that treat components in isolation, our system emphasizes integration and operational efficiency for real-world deployment.

\section{Methodology}

\subsection{System Architecture}

\begin{figure*}[t]
  \centering
  \includegraphics[width=0.9\linewidth]{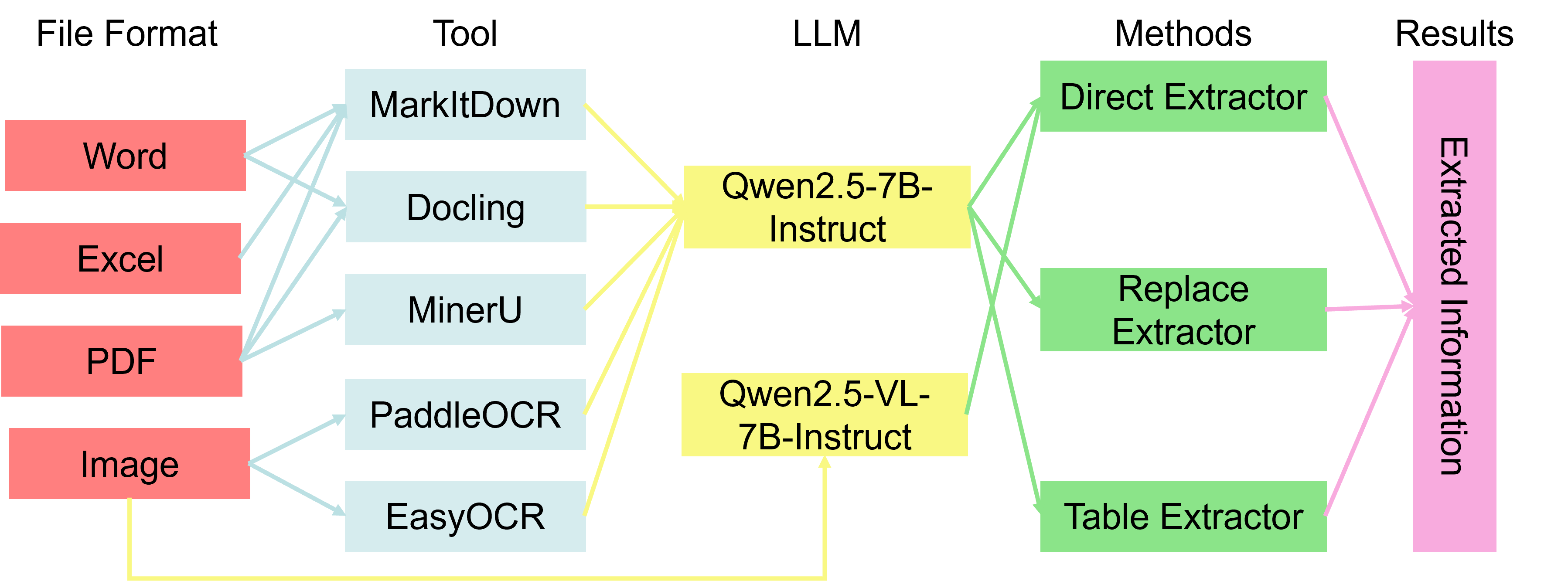}
  \caption {Architecture of the extraction framework showing the two main components: OCR processing, and LLM-based extraction.}
  \label{fig:framework}
\end{figure*}

Our extraction framework comprises two primary components: text extraction and LLM-based information extraction. Figure~\ref{fig:framework} presents an overview of the complete pipeline. Initially, various tools are employed to extract textual content from heterogeneous file formats, including Markdown, Word, Excel, PDF, and images. The extracted text is then passed to a large language model (LLM) to perform task-specific information extraction.

\begin{figure*}[t]
  \centering
  \includegraphics[width=0.9\linewidth]{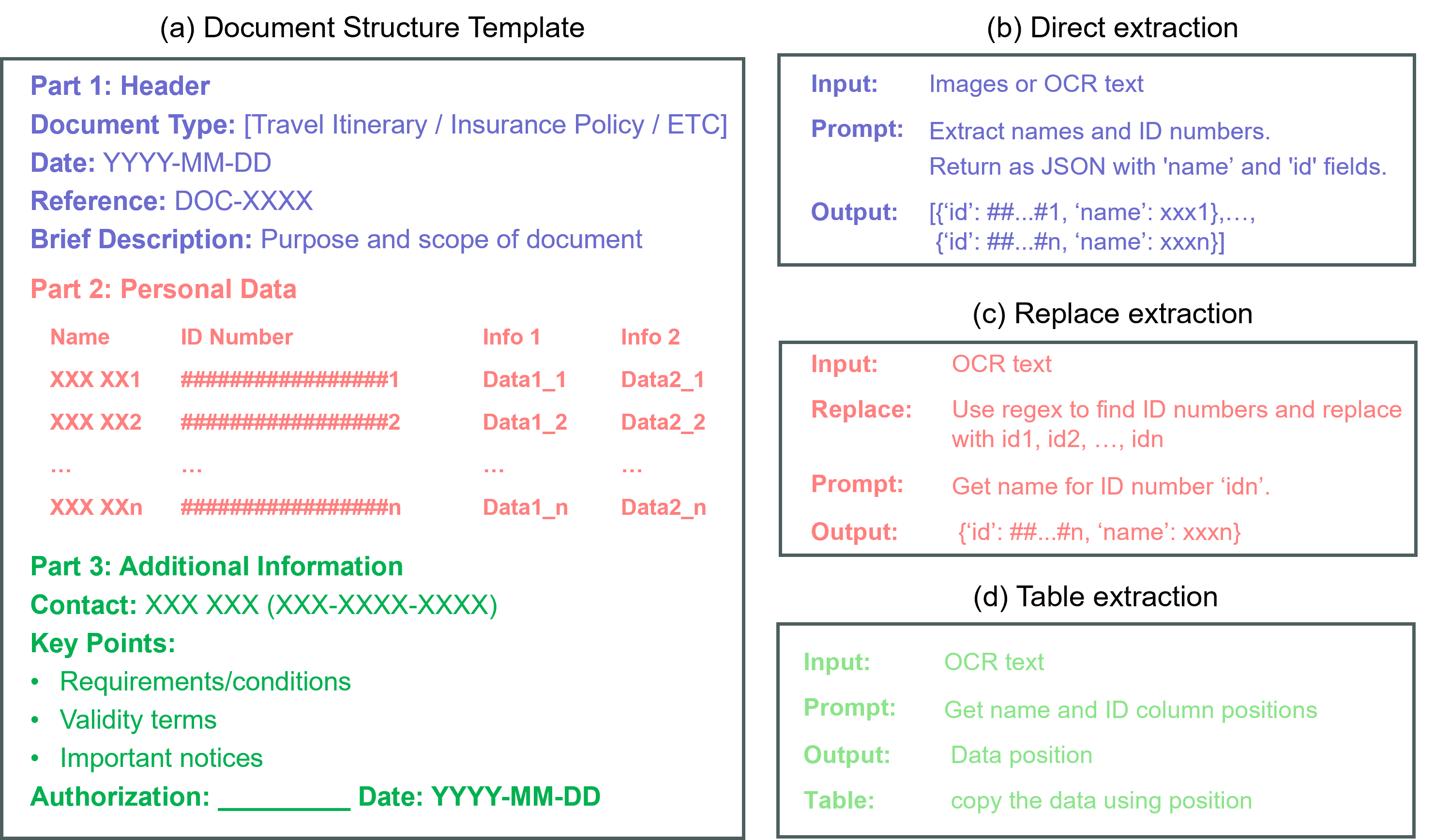}
  \caption {Extraction methods for copy-heavy tasks: (a) Document Structure Template; (b) Direct extraction; (c) Replace extraction; (d) Table extraction.}
  \label{fig:methods}
\end{figure*}

To support different document structures and information needs, we implement multiple extraction strategies. Specifically, our methods fall into three categories, as shown in Figure~\ref{fig:methods}: direct extraction, replacement-based extraction, and table-based extraction, which will be described in detail later. Additionally, we incorporate a multimodal model to directly extract information from images, serving as a baseline for comparison with text-based pipelines.

\subsection{Text Extraction Tools}

The quality of information extraction heavily depends on the accuracy and completeness of upstream text extraction. To support diverse input formats, we integrate a suite of specialized extraction tools, each selected based on its ability to handle specific file types and structural characteristics.

\textbf{MarkItDown} serves as a general-purpose parser for structured text formats, supporting Markdown, Word, Excel, and standard PDF documents. It preserves key structural features such as headings, tables, and semantic markers, enabling clean and semantically meaningful text extraction across multiple formats.

\textbf{Docling} is applied to both Word and PDF files, offering enhanced layout analysis and document hierarchy preservation. It is particularly effective in maintaining reading order and spatial layout, which are essential for downstream structure-aware tasks.

\textbf{MinerU} complements Docling by targeting complex PDF layouts, including multi-column formatting, dense tables, and embedded formulas. It provides fine-grained spatial structure recovery and is optimized for documents where layout fidelity is critical.

\textbf{PaddleOCR} and \textbf{EasyOCR} are employed for image-based documents. PaddleOCR is used as the primary engine due to its high accuracy in multilingual printed text, while EasyOCR acts as a fallback for cases involving handwritten content or degraded image quality.

The use of multiple tools for overlapping file types is intentional. Our goal is to systematically evaluate different extraction solutions on the same document type in order to identify the most effective extraction strategy. Evaluation focuses on two key dimensions: (1) the accuracy of extracted textual content, and (2) the fidelity of spatial and structural information preservation, which is crucial for position-aware downstream tasks such as table extraction or form understanding.

\subsection{Target Information Extraction Methods}

Once the raw text is extracted, we apply LLM-based methods to retrieve the target information. To accommodate different document layouts and content patterns, we design three complementary extraction strategies, each tailored to specific structural characteristics.

\textbf{Direct extraction.} This method applies LLMs or VLMs to perform end-to-end information extraction. In the text-based variant, raw text obtained from upstream extraction tools is passed to an LLM along with task-specific prompts. The vision-based variant bypasses OCR and directly uses document images as input to multimodal models, enabling better spatial understanding at the cost of higher inference overhead.

\textbf{Replace extraction.} To address the challenges of repetitive pattern documents, we adopt a two-step approach. First, structured elements, such as identifiers, are replaced with unique placeholders. The LLM is then prompted to retrieve associated fields based on these placeholders. This design improves consistency, reduces ambiguity, and allows efficient batch processing through prompt reuse. 

\textbf{Table extraction.} For documents with tabular layouts, we combine LLM-based structure recognition with deterministic parsing. The LLM identifies table regions and target cell coordinates, while a rule-based parser extracts the content. This approach minimizes hallucination and reduces generation cost by limiting model output to positional metadata.

To determine the optimal strategy for different document types, we systematically compare the performance of these methods when paired with various text extraction tools.  This enables us to identify the most effective combination in terms of both content accuracy and time consumption, ensuring adaptability and robustness in real-world document scenarios.

\subsection{Model Selection}

We adopt \textbf{Qwen2.5-7B} for text-based extraction and \textbf{Qwen2.5-VL-7B} for vision-language tasks, based on their strong balance between performance and efficiency. The 7B scale offers competitive extraction accuracy with significantly lower inference cost compared to larger models, enabling practical deployment on a single GPU.

Qwen2.5 models also provide robust multilingual support—particularly for Chinese, Japanese, and Korean—and share a unified architecture across text and vision variants, simplifying prompt design and enabling consistent evaluation across modalities. Their strong instruction-following ability ensures reliable structured outputs (e.g., JSON, placeholders), reducing post-processing overhead. These capabilities make them well-suited for our production-oriented information extraction pipeline.

\section{Experiments}

\subsection{Dataset}

To evaluate our multi-method extraction framework, we construct a large-scale synthetic dataset of Chinese identity documents using GPT-4. The dataset simulates real-world diversity in format and content while preserving perfect ground truth.

\textbf{Data Generation Pipeline.} We generate 10--30 identity entries per document using the Faker library (zh\_CN locale), producing realistic Chinese names and 18-digit ID numbers. These entries are embedded into semantically plausible documents (e.g., insurance forms, travel records, registration sheets) via GPT-4 prompts. Each document contains contextual information such as dates, headers, and auxiliary fields.

\textbf{Document Formats.} To evaluate cross-modality robustness, each generated Markdown document is converted into four formats:
\begin{itemize}[itemsep=0pt,parsep=0pt,topsep=0pt]
    \item \textbf{PNG} (100): Rendered HTML to image (via \texttt{imgkit}), simulating scanned documents
    \item \textbf{DOCX} (100): Word documents with tables and formatted sections
    \item \textbf{PDF} (100): Generated using \texttt{WeasyPrint}, preserving layout and structure
    \item \textbf{XLSX} (100): Spreadsheets with tabular identity data
\end{itemize}

The final dataset comprises 400 documents with over 10,000 name-ID pairs.

\subsection{Evaluation Metrics}

We evaluate extraction performance at the (name, ID number) pair level.

\textbf{Accuracy Metrics.} 
\begin{itemize}[itemsep=0pt,parsep=0pt,topsep=0pt]
\item \textbf{Precision:} Ratio of correctly extracted pairs to all extracted pairs \(exact match required\)
\item \textbf{Recall:} Ratio of correctly extracted pairs to total ground truth pairs
\item \textbf{F1 Score:} Harmonic mean of precision and recall
\end{itemize}

\textbf{Efficiency Metrics.} 
\begin{itemize}[itemsep=0pt,parsep=0pt,topsep=0pt]
\item \textbf{Text Extraction Time:} Time for text extraction using each tool (MarkItDown, PaddleOCR, EasyOCR, etc.)
\item \textbf{LLM Time:} Inference time for the extraction model (excluding text extraction)
\item \textbf{Total Time:} End-to-end latency from input to structured output
\end{itemize}

\textbf{Robustness Metrics.} 

\begin{itemize}[itemsep=0pt,parsep=0pt,topsep=0pt]
\item \textbf{Success Rate:} Percentage of documents processed without fatal errors
\item \textbf{Per-Format Accuracy:} Extraction metrics broken down by document type (PNG, DOCX, etc.)
\end{itemize}

\subsection{Results}

We evaluate our multi-method extraction framework on a corpus of 400 synthetic Chinese identity documents spanning four formats (PNG, DOCX, XLSX, PDF), utilizing three extraction paradigms across 16 OCR-LLM configurations. Our findings underscore substantial variation in performance across both extraction strategies and document formats, with method efficacy closely aligned with the structural characteristics of each format.

\textbf{Experimental Overview.} The evaluation encompasses 16 extraction methods categorized into three distinct paradigms: (1) \emph{Direct extraction}, which leverages document parsers or multimodal models; (2) \emph{Replace-based extraction}, which employs rule-based template matching; and (3) \emph{Table-based extraction}, which exploits spatial structure for field localization. Each method was evaluated on up to 100 documents per supported format, resulting in 2,500 test instances. We report precision, recall, F1 score, and processing latency, further decomposed into OCR and LLM inference time.

Table \ref{tab:format_performance} summarizes the performance of the best and worst-performing methods for each document type. For structured formats (DOCX/XLSX), table-based methods achieve perfect $F_1$ scores (1.0), with docling\_table and markitdown\_table delivering 100\% success rates while maintaining minimal latency (0.3--0.5s). In stark contrast, replace-based methods perform poorly on these formats, with markitdown\_replace achieving only $F_1$=0.969 for both DOCX and XLSX, representing the worst performance with perfect rates dropping to 59\% and 54\% respectively. For image-based formats (PNG), the multimodal approach outperforms all OCR-based methods with an $F_1$ score of 0.999, while easyocr\_table catastrophically fails with $F_1$=0.000 and 0\% success rate despite minimal processing time (1.5s). PDF extraction exhibits the most extreme performance variance: docling\_table maintains perfect accuracy ($F_1$=1.0) with low latency (1.6s), whereas mineru\_replace completely fails ($F_1$=0.000, 0\% success rate), highlighting severe method-format incompatibilities. Direct extraction methods consistently incur high LLM inference latency (13.4--13.6s), accounting for over 90\% of total processing time, while table-based approaches achieve 40--50$\times$ speedup through minimal LLM usage.

\begin{table*}
\centering
\small
\begin{tabular}{llccccccccc}
\hline
\textbf{Format} & \textbf{Method} & \textbf{Prec.} & \textbf{Rec.} & \textbf{F1} & \textbf{Succ.} & \textbf{Perf.} & \textbf{OCR} & \textbf{LLM} & \textbf{Total} \\
& & & & & \textbf{Rate} & \textbf{Rate} & \textbf{(s)} & \textbf{(s)} & \textbf{(s)} \\
\hline
\multirow{4}{*}{PNG} & \textcolor{green}{\textbf{multimodal}} & \textcolor{green}{\textbf{.999}} & \textcolor{green}{\textbf{.999}} & \textcolor{green}{\textbf{.999}} & \textcolor{green}{\textbf{100\%}} & \textcolor{green}{\textbf{97\%}} & \textcolor{green}{\textbf{—}} & \textcolor{green}{\textbf{—}} & \textcolor{green}{\textbf{33.9}} \\
 & paddleocr\_table & .998 & .997 & .997 & 100\% & 93\% & 0.3 & 0.3 & 0.6 \\
 & paddleocr\_direct & .998 & .996 & .997 & 100\% & 92\% & 0.4 & 13.4 & 13.8 \\
 & \textcolor{red}{\textbf{easyocr\_table}} & \textcolor{red}{\textbf{.000}} & \textcolor{red}{\textbf{.000}} & \textcolor{red}{\textbf{.000}} & \textcolor{red}{\textbf{0\%}} & \textcolor{red}{\textbf{0\%}} & \textcolor{red}{\textbf{1.2}} & \textcolor{red}{\textbf{0.3}} & \textcolor{red}{\textbf{1.5}} \\
\hline
\multirow{5}{*}{DOCX} & \textcolor{green}{\textbf{docling\_table}} & \textcolor{green}{\textbf{1.00}} & \textcolor{green}{\textbf{1.00}} & \textcolor{green}{\textbf{1.00}} & \textcolor{green}{\textbf{100\%}} & \textcolor{green}{\textbf{100\%}} & \textcolor{green}{\textbf{0.1}} & \textcolor{green}{\textbf{0.3}} & \textcolor{green}{\textbf{0.3}} \\
 & markitdown\_table & 1.00 & 1.00 & 1.00 & 100\% & 100\% & 0.2 & 0.3 & 0.5 \\
 & docling\_direct & 1.00 & 1.00 & 1.00 & 100\% & 99\% & 0.1 & 13.6 & 13.7 \\
 & markitdown\_direct & 1.00 & .999 & 1.00 & 100\% & 98\% & 0.2 & 13.6 & 13.8 \\
 & \textcolor{red}{\textbf{markitdown\_replace}} & \textcolor{red}{\textbf{.969}} & \textcolor{red}{\textbf{.969}} & \textcolor{red}{\textbf{.969}} & \textcolor{red}{\textbf{100\%}} & \textcolor{red}{\textbf{59\%}} & \textcolor{red}{\textbf{0.2}} & \textcolor{red}{\textbf{0.5}} & \textcolor{red}{\textbf{0.7}} \\
\hline
\multirow{3}{*}{XLSX} & \textcolor{green}{\textbf{markitdown\_table}} & \textcolor{green}{\textbf{1.00}} & \textcolor{green}{\textbf{1.00}} & \textcolor{green}{\textbf{1.00}} & \textcolor{green}{\textbf{100\%}} & \textcolor{green}{\textbf{100\%}} & \textcolor{green}{\textbf{0.0}} & \textcolor{green}{\textbf{0.3}} & \textcolor{green}{\textbf{0.3}} \\
 & markitdown\_direct & 1.00 & 1.00 & 1.00 & 100\% & 99\% & 0.0 & 13.5 & 13.5 \\
 & \textcolor{red}{\textbf{markitdown\_replace}} & \textcolor{red}{\textbf{.969}} & \textcolor{red}{\textbf{.969}} & \textcolor{red}{\textbf{.969}} & \textcolor{red}{\textbf{100\%}} & \textcolor{red}{\textbf{54\%}} & \textcolor{red}{\textbf{0.0}} & \textcolor{red}{\textbf{0.5}} & \textcolor{red}{\textbf{0.5}} \\
\hline
\multirow{4}{*}{PDF} & \textcolor{green}{\textbf{docling\_table}} & \textcolor{green}{\textbf{1.00}} & \textcolor{green}{\textbf{1.00}} & \textcolor{green}{\textbf{1.00}} & \textcolor{green}{\textbf{100\%}} & \textcolor{green}{\textbf{100\%}} & \textcolor{green}{\textbf{1.3}} & \textcolor{green}{\textbf{0.3}} & \textcolor{green}{\textbf{1.6}} \\
 & docling\_direct & 1.00 & 1.00 & 1.00 & 100\% & 99\% & 1.3 & 13.5 & 14.9 \\
 & mineru\_direct & 1.00 & 1.00 & 1.00 & 100\% & 99\% & 1.6 & 13.4 & 15.0 \\
 & \textcolor{red}{\textbf{mineru\_replace}} & \textcolor{red}{\textbf{.000}} & \textcolor{red}{\textbf{.000}} & \textcolor{red}{\textbf{.000}} & \textcolor{red}{\textbf{0\%}} & \textcolor{red}{\textbf{0\%}} & \textcolor{red}{\textbf{1.5}} & \textcolor{red}{\textbf{0.0}} & \textcolor{red}{\textbf{1.5}} \\
\hline
\end{tabular}
\caption{Comparative performance metrics showing best (green) and worst (red) extraction methods by document format. The multimodal method uses Qwen2.5-VL-7B while all other methods use Qwen2.5-7B. Prec. = Precision, Rec. = Recall, Succ. Rate = Success Rate (percentage of successful extractions), Perf. Rate = Perfect Rate (percentage achieving perfect F1 score of 1.0). OCR time represents character recognition overhead; LLM time indicates inference latency.}
\label{tab:format_performance}
\end{table*}

\begin{figure*}[t]
  \centering
  \includegraphics[width=0.9\linewidth]{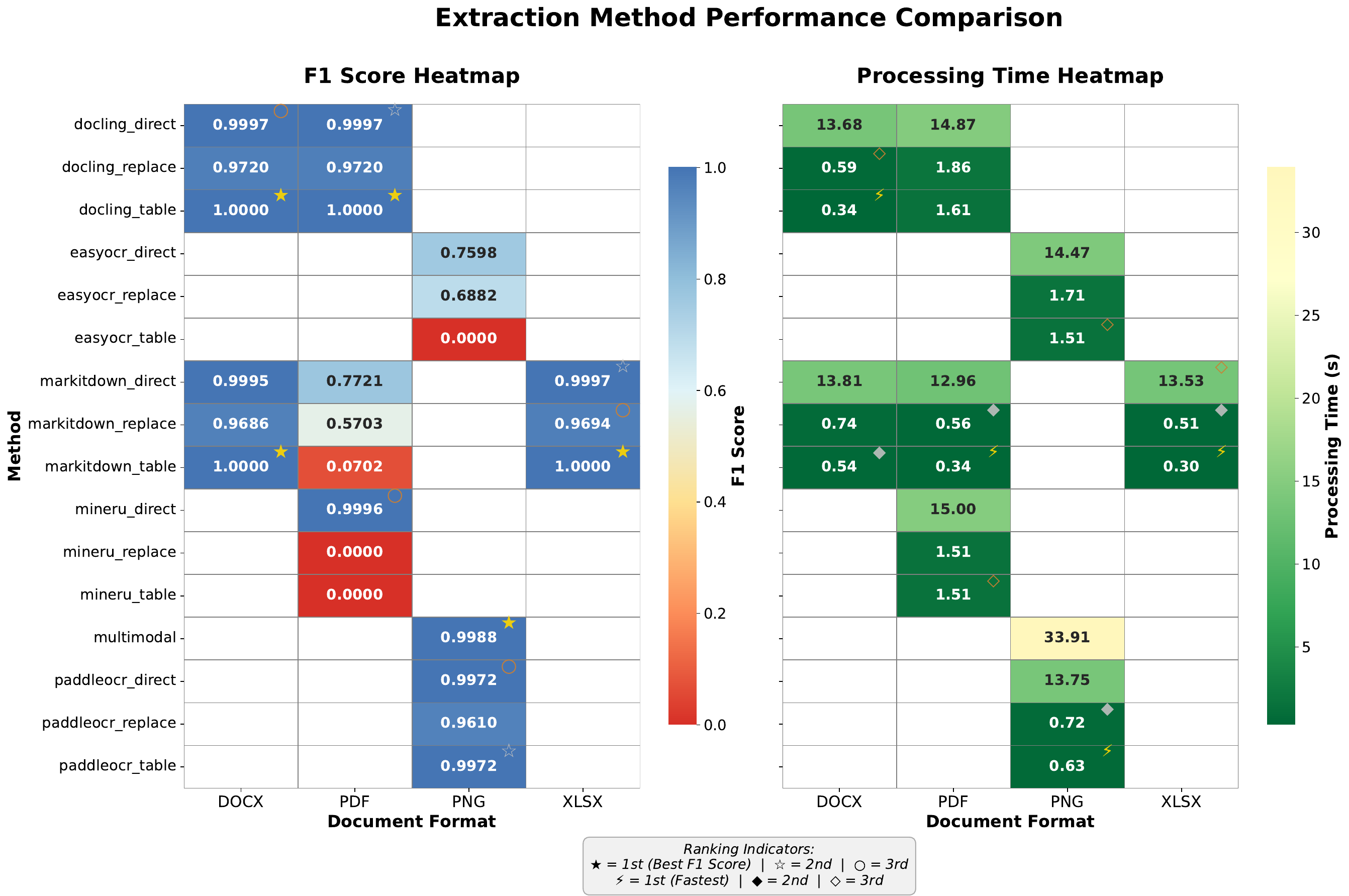}
  \caption{Performance comparison of extraction methods across document formats. The heatmap displays $F_1$ scores and processing time for 16 extraction methods (rows) tested on four document formats (columns). Empty cells denote unsupported format-method combinations. The multimodal method employs Qwen2.5-VL-7B, while all other methods utilize Qwen2.5-7B.}
  \label{fig:heatmap}
\end{figure*}

Figure~\ref{fig:heatmap} provides a holistic view of performance across all 16 methods and formats via dual heatmaps. The $F_1$ score heatmap reveals distinct performance patterns: table-based methods exhibit binary characteristics---either achieving near-perfect extraction ($F_1 \approx 1.0$) or complete failure ($F_1 = 0.0$), indicating strong format dependency. Direct extraction methods demonstrate more consistent but suboptimal performance across formats ($F_1 \approx 0.77$--$0.99$), while replace-based methods show the highest variability, ranging from moderate success to total failure. The processing time heatmap complements these findings, showing that methods with perfect $F_1$ scores often achieve the fastest processing times (0.3--1.6s for table-based approaches), while direct methods consistently require 13--15s due to LLM overhead. The multimodal method presents an outlier with 33.9s processing time for PNG files, trading computational efficiency for extraction accuracy. Empty cells in both heatmaps denote unsupported format-method combinations, particularly evident for specialized parsers like mineru (PDF-only) and OCR-based methods (image formats only). The visualization conclusively demonstrates that no single method achieves universal optimality across all formats, necessitating format-specific extraction strategies for optimal performance.

\begin{figure*}[t]
  \centering
  \includegraphics[width=0.9\linewidth]{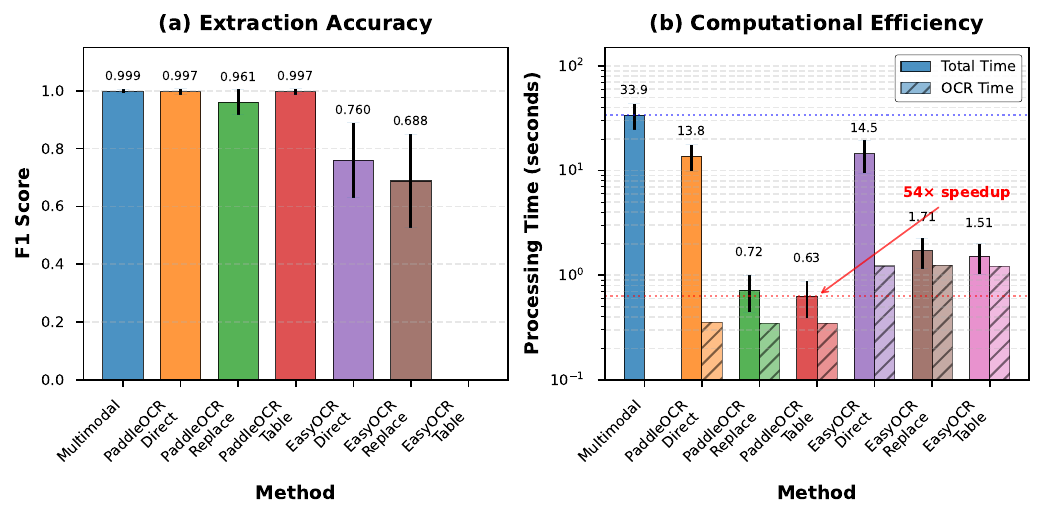}
  \caption{Performance on PNG-based documents: (a) $F_1$ score with standard deviation. (b) Latency comparison across pipelines.}
  \label{fig:png_extraction_performance}
\end{figure*}

\textbf{Image-based Documents (PNG).} For scanned documents with identity cards, we observe stark contrasts between multimodal and OCR-based pipelines, as illustrated in Figure~\ref{fig:png_extraction_performance}. The multimodal vision-language model demonstrates exceptional accuracy ($F_1 = 0.999 \pm 0.007$) through direct visual feature processing, effectively bypassing the cascading errors inherent in sequential OCR pipelines. However, this end-to-end approach incurs prohibitive computational costs, requiring $33.91 \pm 9.49$ seconds per document due to high-dimensional visual encoding and transformer-based attention mechanisms. Among OCR-based methods, we observe significant performance variations based on both the OCR engine and extraction strategy. PaddleOCR consistently outperforms EasyOCR across all extraction paradigms, achieving $F_1$ scores of 0.997, 0.961, and 0.997 for direct, replacement, and table-based extraction respectively, while EasyOCR exhibits substantially degraded performance ($F_1 = 0.760$, $0.688$, and $0.000$). This performance disparity stems fundamentally from EasyOCR's inability to preserve document spatial structure during text extraction, which disrupts the positional relationships critical for accurate field identification in structured documents like identity cards.

The spatial information loss particularly impacts table-based extraction, where EasyOCR completely fails ($F_1 = 0.000$) as the LLM cannot generate valid coordinates without reliable spatial encoding. In contrast, PaddleOCR maintains precise spatial mappings, enabling the table-based extraction strategy to achieve optimal performance at $0.63 \pm 0.24$ seconds—a $54\times$ speedup over the multimodal baseline—while preserving near-perfect accuracy ($F_1 = 0.997 \pm 0.011$). This method minimizes LLM inference to coordinate generation only, leveraging the preserved document structure to reduce computational overhead from $\sim$34 seconds to sub-second latency. The replacement method with PaddleOCR, employing regex patterns for standardized fields and batch processing for names, offers a middle ground at $0.72 \pm 0.28$ seconds ($F_1 = 0.961 \pm 0.044$), while direct extraction requires $13.75 \pm 3.87$ seconds despite high accuracy ($F_1 = 0.997 \pm 0.010$) due to processing entire OCR outputs. The PaddleOCR table-based approach thus emerges as the optimal solution for production deployment, combining superior spatial preservation with minimal LLM computation. The marginal 0.2\% accuracy trade-off compared to multimodal methods represents an acceptable compromise given the 54-fold efficiency gain crucial for large-scale document processing systems.

\begin{figure*}[t]
  \centering
  \includegraphics[width=0.9\linewidth]{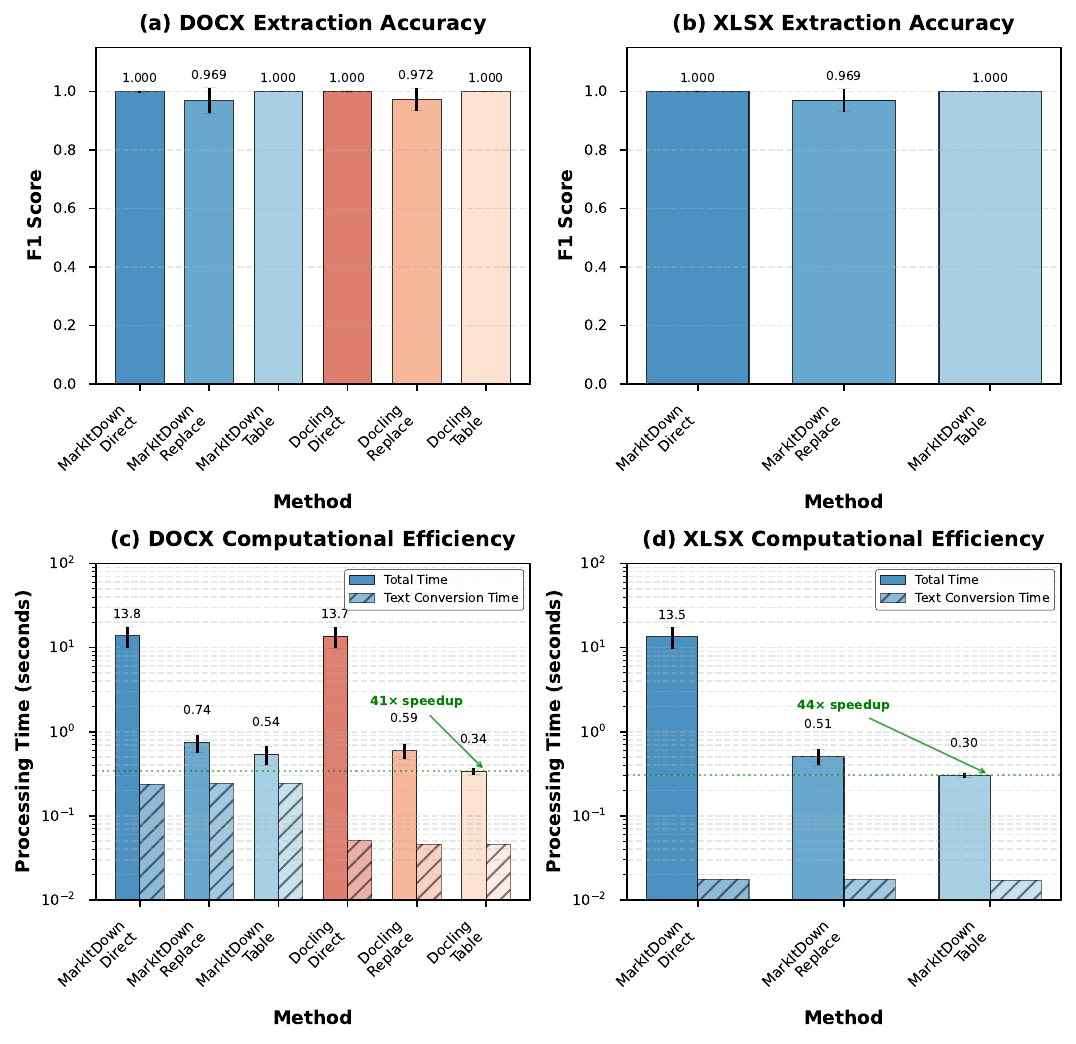}
  \caption{Performance comparison of different extraction methods for native office documents processing. (a) DOCX extraction accuracy. (b) XLSX extraction accuracy. (c) DOCX computational efficiency. (d) XLSX computational efficiency.}
  \label{fig:office}
\end{figure*}

\textbf{Structured Office Documents (DOCX/XLSX).} Our evaluation of structured office document extraction reveals fundamentally different performance characteristics compared to image-based formats, as illustrated in Figure~\ref{fig:office}. Native documents exhibit near-perfect accuracy across all extraction methodologies due to their inherent machine-readable structure and preserved spatial information. For DOCX files, both MarkItDown and Docling frameworks successfully extract the document's spatial structure, enabling high-fidelity extraction across all paradigms. Docling's table-based approach achieves optimal performance with perfect accuracy ($F_1 = 1.000$) and exceptional efficiency at $0.34 \pm 0.03$ seconds—a $41\times$ speedup compared to direct extraction ($13.68 \pm 3.85$ seconds)—by leveraging structured table representations that minimize LLM processing to coordinate generation only. MarkItDown's table method follows closely with identical accuracy ($F_1 = 1.000$) at $0.54 \pm 0.13$ seconds, representing a $26\times$ improvement over its direct counterpart.

While replacement strategies offer competitive latency (Docling: $0.59 \pm 0.13$s, MarkItDown: $0.74 \pm 0.18$s), they sacrifice accuracy ($F_1 = 0.972 \pm 0.038$ and $0.969 \pm 0.044$ respectively) due to the LLM's occasional misidentification when matching ID numbers to corresponding names—a limitation stemming from the model's reliance on contextual inference rather than explicit structural cues. This name-matching error could be mitigated through prompt engineering to provide more explicit matching instructions or fine-tuning the general-purpose LLM on document-specific extraction tasks. For XLSX spreadsheets, where only MarkItDown provides support, the inherent tabular structure amplifies these efficiency gains: table-based extraction achieves perfect accuracy ($F_1 = 1.000$) with remarkable $0.30 \pm 0.02$ seconds processing time—a $44\times$ acceleration over direct methods ($13.53 \pm 3.82$ seconds). This dramatic improvement stems from the natural alignment between spreadsheet cell structures and table-based extraction paradigms, eliminating the need for complex content interpretation. The replacement method maintains reasonable efficiency ($0.51 \pm 0.11$s) but exhibits similar accuracy degradation ($F_1 = 0.969 \pm 0.039$) as observed in DOCX processing, with errors predominantly occurring in name-ID association tasks. These results establish table-based extraction as the unequivocally superior approach for native office documents, where the preserved document structure enables both perfect accuracy and minimal computational overhead, making it ideal for enterprise-scale document processing pipelines requiring both precision and throughput.

\textbf{Portable Document Format (PDF).} Our empirical analysis of PDF document processing reveals a fundamental trade-off between computational efficiency and extraction accuracy across different methodological approaches. MarkItDown prioritizes processing speed, achieving minimal OCR preprocessing overhead (0.056s), while Docling and MinerU adopt more computationally intensive strategies with OCR latencies of approximately 1.3--1.5s. However, this efficiency-accuracy trade-off manifests differently across extraction paradigms. MarkItDown's lightweight approach, which employs symbolic markers for spatial structure representation, exhibits severely degraded performance across all extraction strategies: direct extraction ($F_1=0.772$), replacement extraction ($F_1=0.570$), and table-based extraction ($F_1=0.070$).

\begin{figure*}[t]
  \centering
  \includegraphics[width=0.9\linewidth]{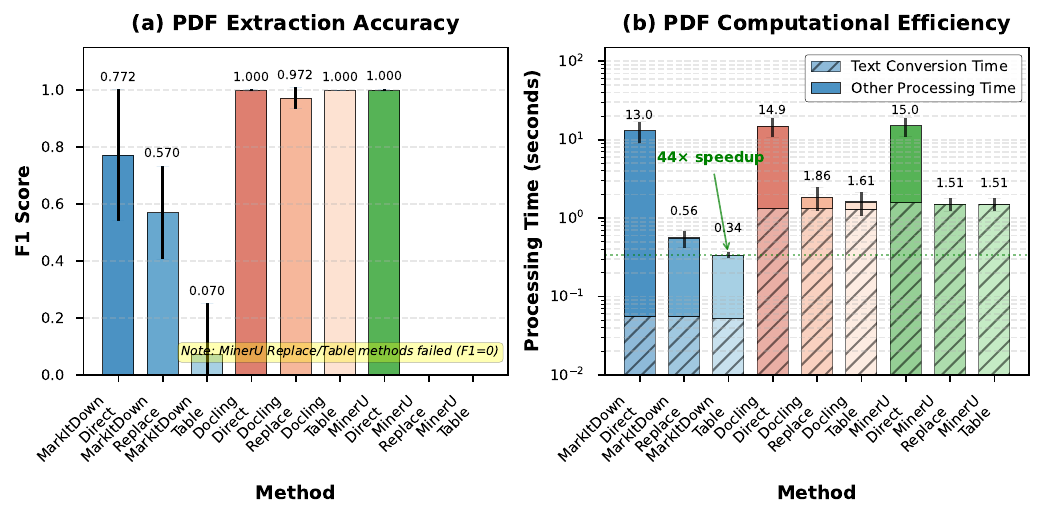}
  \caption{Performance comparison of different extraction methods for portable document format processing. (a) Extraction accuracy measured by F1 score with error bars showing standard deviation. (b) Computational efficiency comparison showing total latency (with error bars) and text conversion time for the markitdown, docling and minerU pipeline.}
  \label{fig:pdf_extraction_performance}
\end{figure*}

The divergence in extraction efficacy between Docling and MinerU, despite their comparable preprocessing costs, underscores the critical importance of spatial representation strategies in document understanding. Docling maintains near-perfect accuracy across all extraction paradigms, with its table-based approach achieving optimal performance ($F_1=1.0$). In stark contrast, MinerU's reliance on HTML-style structural tags (e.g., \texttt{<td>}, \texttt{<tr>}) for encoding spatial relationships proves incompatible with text-based information extraction, resulting in catastrophic failure for replacement and table-based methods ($F_1=0.0$). The framework only maintains competitive performance in direct extraction mode ($F_1=0.9996$), where spatial structure parsing is bypassed entirely. These empirical findings, illustrated in Figure~\ref{fig:pdf_extraction_performance}, establish Docling's table-based methodology as the optimal solution for PDF information extraction, successfully reconciling the competing demands of computational efficiency (1.61s total processing time) and extraction fidelity.

\subsection{Discussion}

Our comprehensive evaluation reveals a fundamental insight: the optimal extraction strategy for copy-heavy documents is intrinsically tied to document modality and structure, challenging the prevailing one-size-fits-all approaches in production systems. The 54× performance differential between table-based methods (0.97s) and multimodal approaches (33.91s) underscores a critical trade-off—while end-to-end models offer superior robustness through direct visual understanding, their computational overhead remains prohibitive for high-throughput scenarios characteristic of copy-heavy tasks. This finding suggests that the field's pursuit of universal extraction models may be misguided for repetitive document processing, where structural priors can be effectively exploited.

The stark performance dichotomy between OCR engines (PaddleOCR F1=0.997 vs. EasyOCR F1=0.000 for table extraction) reveals that spatial structure preservation, rather than character recognition accuracy alone, determines extraction success. This challenges conventional OCR evaluation metrics and highlights the need for structure-aware benchmarks. Moreover, the consistent superiority of table-based methods across native formats (100\% accuracy at <1s) demonstrates that explicit structural modeling outperforms both semantic understanding (direct extraction) and pattern matching (replacement extraction) when document layouts are predictable—a defining characteristic of copy-heavy scenarios.

From a production standpoint, our results advocate for adaptive, format-aware architectures over monolithic solutions. The multimodal approach's resilience to OCR failures positions it as an ideal fallback mechanism, while table-based extraction serves as the workhorse for structured documents. This hierarchical strategy—rapid format detection followed by method-specific routing—enables systems to process heterogeneous document streams at scale without sacrificing accuracy. The framework thus provides a blueprint for reconciling the competing demands of accuracy, efficiency, and robustness in real-world document processing pipelines.

\section{Conclusion}

We present a systematic framework for information extraction from copy-heavy documents, demonstrating that the repetitive nature of such tasks—rather than being a mere computational burden—can be strategically exploited through intelligent method selection. Our comprehensive evaluation of 25 method combinations across diverse documents reveals that optimal extraction strategies must align with document characteristics: table-based methods for structured formats (achieving perfect accuracy at sub-second latency), multimodal approaches for degraded images (F1=0.999), and adaptive routing for heterogeneous streams.

The core contribution lies not in proposing novel extraction techniques, but in establishing a principled approach to method selection for copy-heavy scenarios. By recognizing that document structure dictates optimal extraction paradigms, we achieve a 54× speedup while maintaining accuracy—a critical requirement for enterprise-scale deployment. This work challenges the field's emphasis on universal models, showing that domain-specific solutions remain indispensable when processing millions of similar documents daily.

For practitioners, our framework offers immediate value: implement table-based extraction as the default for structured documents, maintain multimodal models as robust fallbacks, and use format detection for intelligent routing. For researchers, we highlight unexplored opportunities in hybrid architectures that combine the efficiency of structural methods with the robustness of end-to-end models. As document processing increasingly underpins digital transformation, our work provides both theoretical insights and practical tools for building extraction systems that scale without compromising quality.


\bibliography{custom}

\end{document}